\documentclass{article} 

\usepackage{microtype}
\usepackage{graphicx}
\usepackage{subfigure}
\usepackage{supertabular}
\usepackage{booktabs} 

\usepackage[preprint]{neurips2024}

\usepackage{amsmath,amsfonts,bm}

\def\eqref#1{equation~\ref{#1}}

\def\1{\bm{1}}

\DeclareMathAlphabet{\mathsfit}{\encodingdefault}{\sfdefault}{m}{sl}
\SetMathAlphabet{\mathsfit}{bold}{\encodingdefault}{\sfdefault}{bx}{n}

\usepackage{hyperref}
\usepackage{url}
\usepackage{graphicx}
\usepackage{subcaption}
\usepackage{amsmath,amssymb}
\usepackage{mathrsfs}
\usepackage[normalem]{ulem}
\usepackage{booktabs}
\usepackage{verbatim}
\usepackage{enumitem}
\usepackage{multirow}
\usepackage{xltabular}
\usepackage{siunitx}
\usepackage{soul}
\usepackage{mathtools}
\usepackage{amsthm}

\usepackage[capitalize,noabbrev]{cleveref}

\theoremstyle{plain}

\theoremstyle{definition}

\theoremstyle{remark}

\usepackage[textsize=tiny]{todonotes}

\DeclareMathOperator{\rightN}{right}
\DeclareMathOperator{\leftN}{left}

\usepackage{xspace}
\makeatletter
\DeclareRobustCommand\onedot{\futurelet\@let@token\@onedot}
\newcommand{\@onedot}{\ifx\@let@token.\else.\null\fi\xspace}
\makeatother

\newcommand{\ie}{i.\,e.,\xspace}
\newcommand{\eg}{e.\,g.,\xspace}

\author{Alexander Mattick and
        Christopher Mutschler \\
       Fraunhofer Institute for Integrated Circuits (IIS), Fraunhofer IIS \\
       Nuremberg, Germany\\
       \texttt {\{FirstName.LastName\}@iis.fraunhofer.de} } 
\title{Reinforcement Learning for Node Selection in Branch-and-Bound}

\begin{document}







\maketitle

\begin{abstract}
A big challenge in branch and bound lies in identifying the optimal node within the search tree from which to proceed. 
Current state-of-the-art selectors utilize either hand-crafted ensembles that automatically switch between naive sub-node selectors, or learned node selectors that rely on individual node data.
We propose a novel simulation technique that uses reinforcement learning (RL) while considering the entire tree state, rather than just isolated nodes.
To achieve this, we train a graph neural network that produces a probability distribution based on the path from the model's root to its ``to-be-selected'' leaves. Modelling node-selection as a probability distribution allows us to train the model using state-of-the-art RL techniques that capture both intrinsic node-quality and node-evaluation costs.
Our method induces a high quality node selection policy on a set of varied and complex problem sets, despite only being trained on specially designed, synthetic travelling salesmen problem (TSP) instances.
Using such a fixed pretrained policy shows significant improvements on several benchmarks in optimality gap reductions and per-node efficiency under strict time constraints.
\end{abstract}

\section{Introduction}
\label{sec:Introduction}

The optimization paradigm of mixed integer programming plays a crucial role in addressing a wide range of complex problems, including scheduling~\citep{Bayliss2017ASS}, process planning~\citep{Floudas2005MixedIL}, and network design~\citep{MENON2013690}. 
A prominent algorithmic approach employed to solve these problems is \emph{branch-and-bound (BnB)}, which recursively subdivides the original problem into smaller sub-problems through variable branching and pruning based on inferred problem bounds. BnB is also one of the main algorithms implemented in SCIP~\citep{BestuzhevaEtal2021OO,BestuzhevaEtal2021ZR}, a state-of-the art mixed integer linear and mixed integer nonlinear solver.




An often understudied aspect is the node selection problem, which involves determining which nodes within the search tree are most promising for further exploration. 
This is due to the intrinsic complexity of understanding the emergent effects of node selection on overall performance for human experts. Contemporary methods addressing the node selection problem typically adopt a per-node perspective~\citep{yilmazStudyLearningSearch2021,heLearningSearchBranch,MORRISON201679}, incorporating varying levels of complexity and relying on imitation learning (IL) from existing heuristics~\citep{yilmazStudyLearningSearch2021,heLearningSearchBranch}.
However, they fail to fully capture the rich structural information present within the branch-and-bound tree itself.

\begin{figure*}[t]
    \centering
    \includegraphics[width=\linewidth]{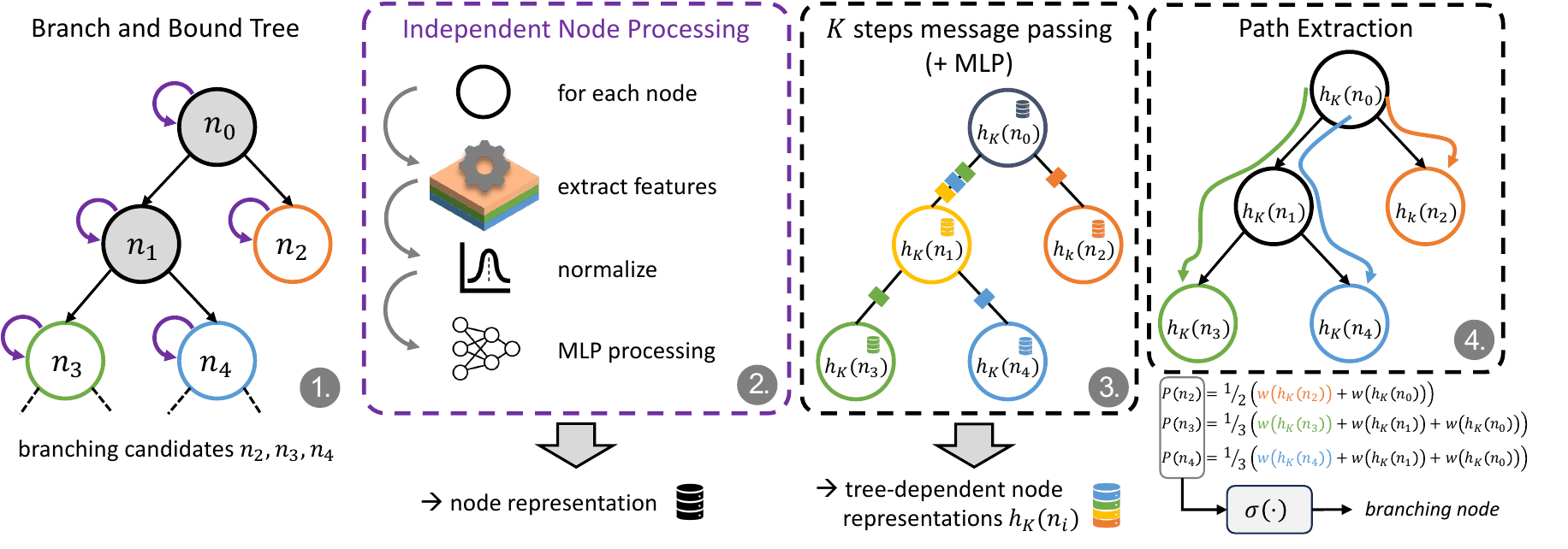}
    \vspace{-6mm}
    \caption{
    Our method:
    (1) SCIP solves individual nodes and executes existing heuristics.
    (2) Features are extracted from every branch-and-bound node and sent to individual normalization and embedding.
    (3) The node embeddings are subject to $K$ steps of GNN message passing on the induced tree-structure.
    (4) Based on the node embeddings, we generate root-to-leave paths, from which we sample the next node.
    The resulting node is submitted to SCIP and we return to step 1.
    }
    \label{fig:MethodOverview}
\end{figure*}

We propose a novel selection heuristic that leverages the power of bi-simulating the branch-and-bound tree with a neural network-based model and that employs reinforcement learning (RL) for heuristic training, see Fig.~\ref{fig:MethodOverview}. 
To do so, we reproduce the SCIP state transitions inside our neural network structure (bi-simulation), which allows us to take advantage of the inherent structures induced by branch-and-bound.
This allows the RL policy to directly account for the BnB tree's dynamics.

We reason that RL specifically is a good fit for this type of training as external parameters outside the pure quality of a node have to be taken into account.
For example, a node $A$ might promise a significantly bigger decrease in the expected optimality gap than a second node $B$, but node $A$ might take twice as long to evaluate, making $B$ the ``correct'' choice despite its lower theoretical utility.
By incorporating the bi-simulation technique, we can effectively capture 
and propagate relevant information throughout the tree. 




\section{Branch and Bound}
\label{sec:Background}

BnB is one of the most effective methods for solving mixed integer programming (MIP) problems. It recursively solves relaxed versions of the original problem, gradually strengthening the constraints until it finds an optimal solution. The first step relaxes the original MIP instance into a tractable subproblem by dropping all integrality constraints such that the subproblem can later be strictified into a MIP solution.
For simplicity, we focus our explanation to the case of mixed integer linear programs (MILP) while our method theoretically works for any type of constraint allowed in SCIP (see nonlinear results in Sec.~\ref{sec:exp:MINLPLIB}, and \citet{bestuzhevaGlobalOptimizationMixedInteger2023}. Concretely a MILP has the form
\begin{equation}
P_{\text{MILP}}=\min \{ c^T_1x + c^T_2y | Ax + By \geq  b, y\in \mathbb{Z}^n\},
\end{equation}
where $c_1$ and $c_2$ are coefficient vectors, $A$ and $B$ are constraint matrices, and $x$ and $y$ are variable vectors. The integrality constraint $y\in \mathbb{Z}^n$ requires $y$ to be an integer. In the relaxation step, this constraint is dropped, leading to the following simplified problem:
\begin{equation}
P_{\text{relaxed}} = \min \{ c^T_1x + c^T_2y | Ax + By \geq  b\}. \label{eq:principleRelax}
\end{equation}
Now, the problem becomes a linear program without integrality constraints, which can be exactly solved using the Simplex~\citep{DANTZIG198243} or other efficient linear programming algorithms.

After solving the relaxed problem, BnB proceeds to the branching step: First, a non-integral $y_i$ is chosen. 
The branching step then derives two problems: The first problem (Eq.~\ref{eq:rel1}) adds a lower bound to variable $y_i$, while the second problem (Eq.~\ref{eq:rel2}) adds an upper bound to variable $y_i$. 
These two directions represent the rounding choices to enforce integrality for $y_i$:\footnote{There are non-binary, ``wide'' branching strategies which we will not consider here explicitly. However, our approach is flexible enough to allow for arbitrary branching width. See~\citet{MORRISON201679} for an overview.}
\begin{align}
    P_{\text{relaxed}}^1 &= \min \{ c^T_1x + c^T_2y | Ax + By \geq  b, y_i\leq \lfloor c\rfloor\}\label{eq:rel1}\\
    P_{\text{relaxed}}^2 &= \min \{ c^T_1x + c^T_2y | Ax + By \geq  b, y_i\geq \lceil c\rceil\}\label{eq:rel2}
\end{align}
The resulting decision tree, with two nodes representing the derived problems can now be processed recursively. However, a naive recursive approach exhaustively enumerates all integral vertices, leading to an impractical computational effort. Hence, in the bounding step, nodes that are deemed worse than the currently known best solution are discarded.
To do this, BnB stores previously found solutions which can be used as a lower bound to possible solutions. 
If a node has an upper bound larger than a currently found integral solution, no node in that subtree has to be processed.

The interplay of these three steps—relaxation, branching, and bounding—forms the core of branch-and-bound. It enables the systematic exploration of the solution space while efficiently pruning unpromising regions. Through this iterative process, the algorithm converges towards the globally optimal solution for the original MIP problem, while producing optimality bounds at every iteration.

\section{Related Work}
\label{sec:RelatedWork}

While variable selection through learned heuristics has been studied a lot (see e.g.~\citet{Parsonson2022ReinforcementLF} or~\citet{Etheve2020ReinforcementLF}), learning node selection, where learned heuristics pick the best node to continue, have only rarely been addressed in research.
We study learning algorithms for node selection in state-of-the-art branch-and-cut solvers.
Prior work that learns such node selection strategies made significant contributions to improve the efficiency and effectiveness of the optimization.

Notably, many approaches rely on per-node features and Imitation Learning (IL). \citet{ottenCaseStudyComplexity} examined the estimation of subproblem complexity as a means to enhance parallelization efficiency. By estimating the complexity of subproblems, the algorithm can allocate computational resources more effectively. \citet{yilmazStudyLearningSearch2021} employed IL to directly select the most promising node for exploration. Their approach utilized a set of per-node features to train a model that can accurately determine which node to choose. \citet{heLearningSearchBranch} employed support vector machines and IL to create a hybrid heuristic based on existing heuristics. By leveraging per-node features, their approach aimed to improve node selection decisions. While these prior approaches have yielded valuable insights, they are inherently limited by their focus on per-node features. 

\citet{labassiLearningCompareNodes} proposed the use of Siamese graph neural networks, representing each node as a graph that connects variables with the relevant constraints. Their objective was direct imitation of an optimal diving oracle. This approach facilitated learning from node comparisons and enabled the model to make informed decisions during node selection. However, the relative quality of two nodes cannot be fully utilized to make informed branching decisions as interactions between nodes remain minimal (as they are only communicated through a final score and a singular global embedding containing the primal and dual estimate). While the relative quality of nodes against each other is used, the potential performance is limited as the overall (non-leaf) tree structure is not considered. Existing methods are further limited by their heavy reliance on pure IL that will be constrained by the performance of the existing heuristics. Hence, integrating RL to strategically select nodes holds great promise.

\section{Methodology}
\label{sec:Method}

We model the node selection process as a sequential Markov Decision Process (MDP).
An MDP consists of a set of states $s\in\mathcal{S}$, actions $a\in \mathcal{A}$, reward $R(s)$ and a (typically unknown) transition distribution $T:\mathcal{S}\times\mathcal{A}\to\mathcal{S}$. We train a decision policy $\pi:\mathcal{S}\to\mathcal{A}$ that selects actions in states. Our objective is to find the policy $\pi^*$, which maximizes the expected discounted sum of future rewards:
\begin{equation}
    \pi^* =\operatorname{argmax}_\pi \mathbb{E}\left[\sum^{t_{\text{max}}}_{t=0} \gamma^t R(s_t)\pi(a_t|s_t)T(s_{t+1}|s_t,a_t)\right],
\end{equation}
where $0<\gamma<1$ is a factor to trade immediate over future rewards.~\citep{sutton2018reinforcement}

RL is the process of solving such MDPs by repeatedly rolling out $\pi$ on an environment and using the resulting trajectory to maximize $\pi$.
Several solvers for MDPs exist, such as DQN \citep{Mnih2015} or PPO \citep{Schulman2017ProximalPO}. In this paper we use PPO to solve the MDP.

In our case, we phrase our optimization problem as an MDP, where states $s_t$ are represented using directed branch-and-bound trees, actions $a_t$ are all selectable leaf-nodes, and the reward $R$ is set according to Eq.~\ref{eq:reward} in Sec.~\ref{sec:RewardDefinition}.
The transitions are given by the SCIP-solver and its heuristics.
To solve the MDP, PPO needs a representation for the policy $\pi$ and a model describing the state-value-function, defined as $V(s) = \max_a Q(s_t,a)$ (see \citet{Schulman2017ProximalPO}).

In our model, we represent the branch-and-bound tree as a directed graph neural network whose nodes and edges are in an 1:1 correspondence to the branch-and-bound nodes and edges.
One benefit of this is that the resulting directed graph automatically grows/shrinks the action space as new nodes are added/pruned from the tree, since the policy's action-space is directly parameterized by the number of selectable leaves.
This allows our network to phrase node selection as an explicit distributional problem, rather than the implicit comparison based one used by \citet{labassiLearningCompareNodes}.

\subsection{Reward definition}\label{sec:RewardDefinition}
A common issue in MIP training is the lack of good reward functions.
For instance, the primal-dual integral used in \citet{Gasse2022TheML} has the issue of depending on the scale of the objective, needing a primally feasible solution at the root, and being very brittle if different instance difficulties exist in the instance pool.
These issues where circumvented by generating similar instances, while also supplying a feasible solution at the root.
However, to increase generality, we do not want to make as strict assumptions as having similar instance difficulties, introducing a need for a new reward function.
A reasonable reward measure captures the achieved performance against a well-known baseline. 
In our case, we choose our reward, to be
\begin{equation}
    r = -\left(\frac{\text{\textit{gap(node selector)}}}{\text{\textit{gap(scip)}}}-1\right),\label{eq:reward}
\end{equation}
which represents the relative performance of our selector over the performance achieved by SCIP.

Intuitively, the aim of this reward function is to decrease the optimality gap achieved by our node-selector (i.e., \textit{gap(node selector)}), normalized by the results achieved by the existing state-of-the-art node-selection methods in the SCIP~\citep{BestuzhevaEtal2021OO} solver (i.e., \textit{gap(scip)}).
This accounts for varying instance hardness in the dataset.
Further, we shift this performance measure, such that any value $>0$ corresponds to our selector being superior to existing solvers, while a value $<0$ corresponds to our selector being worse, and clip the reward between $(1,-1)$ to ensure symmetry around zero as is commonly done in prior work (see, e.g. \citet{Mnih2015,DAYAL2022109241}).
This formulation has the advantage of looking at relative improvements over a baseline, rather than absolute performance, which allows us to use instances of varying complexity during training.


\subsection{Tree Representation}
\label{subsec:MethodRepresentation}

To represent the MDP's state, we propose a novel approach that involves bi-simulating the existing branch-and-bound tree using a graph neural network (GNN). 
Specifically, we consider the state to be a directed tree $T=(V,E)$, where $V$ are individual subproblems created by SCIP, and $E$ are edges set according to SCIP's node-expansion rules. More precisely, the tree's root is the principle relaxation (Eq.~\ref{eq:principleRelax}), and its children are the two rounding choices (Eqs.~\ref{eq:rel1} and~\ref{eq:rel2}). BnB proceeds to recursively split nodes, adding the splits as children to the node it was split from.

For processing inside the neural network, we extract features (see Appendix~\ref{sec:features}) from each node, keeping the structure intact: $T=(\operatorname{extract}(V), E)$.
We ensure that the features stay at a constant size, independent from, e.g., the number of variables and constraints, to enable efficient batch-evaluation of the entire tree.

{For processing the tree $T$, we use message-passing from the children to the parent.
Pruned or missing nodes are replaced with a constant to maintain uniformity over the graph structure.
Message-passing enables us to consider the depth-$K$ subtree under every node by running $K$ steps of message passing from the children to the parent.}


Concretely, the internal representation can be thought of initializing $h_0(n) = x(n)$ (with $x(n)$ being the features associated with node $n$) and then running $K$ iterations jointly for all nodes $n$:
\begin{equation}
    h_{t+1}(n) = h_{t}(n) + \operatorname{emb}\left(\frac{h_t(\leftN(n)) + h_t(\rightN(n))}{2}\right),
\end{equation}
where $\leftN(n)$ and $\rightN(n)$ are the left and right children of $n$, respectively, $h_t(n)$ is the hidden representation of node $n$ after $t$ steps of message passing, and $\operatorname{emb}$ is a function that takes the mean hidden state of all embeddings and creates an updated node embedding.
Doing this for $K$ steps aggregates the information of the depth-K limited subtree of $n$ into the node $n$.

\subsection{RL for Node Selection}
\label{subsec:MethodRL}

While the GNN model is appealing, it is impossible to train using contemporary imitation learning techniques, as the expert's action domain (i.e., leaves) may not be the same as the policy's action domain, meaning that the divergence between these policies is undefined.

To solve this problem, we choose to use model-free reinforcement learning techniques to directly maximize the probablity of choosing the correct node.
State-of-the-art reinforcement learning techniques, such as proximal policy optimization (PPO)~\cite{Schulman2017ProximalPO}, need to be able to compute two functions: The value function $V(s) = \max_{a\in \mathcal{A}}Q(s,a)$, and the likelihood function $\pi(a | s)$.
As it turns out, there are efficient ways to compute both of these values from a tree representation:

Firstly, we can produce a probability distribution of node-selections (\ie our actions) by computing the expected ``weight'' across the unique path from the graph's root to the ``to-be-selected'' leaves.
These weights are mathematically equivalent to the unnormalized log-probabilities of choosing a leaf-node by recursively choosing either the right or left child with probability $p$ and $1-p$ respectively.
The full derivation of this can be found in appendix~\ref{sec:theoryNodeSelect}.
Concretely, let $n$ be a leaf node in the set of choosable nodes $\mathscr{C}$, also let $P(r,n)$ be the unique path from the root $r$ to the candidate leaf node, with $|P(r,n)|$ describing its length.
We define the expected path weight $W'(n)$  to a leaf node $n\in\mathscr{C}$ as
\begin{equation}
    W'(n) = \frac{1}{|P(r,n)|} \sum_{u \in P(r,n)} W(h_K(u)).\label{eq:Weight}
\end{equation}
Selection now is performed in accordance to sampling from a selection policy $\pi$ induced by
\begin{equation}
    \pi(n|\operatorname{tree}) = \operatorname{softmax}\left(\{ W'(n) | \forall n\in\mathscr{C}\right\}).\label{eq:policy}
\end{equation}
Intuitively, this means that we select a node exactly if the expected utility along its path is high.
Note that this definition is naturally self-correcting as erroneous over-selection of one subtree will lead to that tree being completed, which removes the leaves from the selection pool $\mathscr{C}$.

{The general value function $V(s)$, can be constructed similarly by using the fact that $V(s) = \max_{a\in \mathcal{A}} Q(s,a)$ and the fact that our actions correspond to individual nodes:}
\begin{align}
    Q(n|s) &= \frac{\tilde Q(n|s)}{|P(r,n)|}\\
    \tilde Q(n|s) &= \tilde Q(\text{left}|s)+ \tilde Q(\text{right}|s) + q(h_K(n)|s)\\
    V(s) &= \left\{ \max Q(n) \hspace{2mm}|\hspace{2mm} \forall n\in \mathscr{C} \right\}
\end{align}
where $q(h_n)$ is the per-node estimator, $\tilde Q$ the unnormalized Q-value, and $\mathscr{C}$ is the set of open nodes as proposed by the branch-and-bound method.
{Since we can compute the action likelihood $\pi(a|s)$, the value function $V(s)$, and the Q-function $Q(s,a)$, we can use any Actor-Critic method (like \cite{Schulman2017ProximalPO,schulmanHighDimensionalContinuousControl2015,Mnih2016AsynchronousMF}) for training this model.
For this paper, we chose to use \citet{Schulman2017ProximalPO}, due to its ease of use and robustness.}

{
According to these definitions, we only need to predict the node embeddings for each node $h_K(n)$, the per-node q-function $q(h_K(n)|s)$, and the weight of each node $W(h_K(u))$.
We parameterize all of these as MLPs (more architectural details app.~\ref{sec:architecture}.
}

This method provides low, but measurable overhead compared to existing node selectors, even if we discount the fact that our Python-based implementation is vastly slower than SCIP's highly optimized C-based implementations.
Hence, we focus our model on being efficient at the beginning of the selection process, where good node selections are exponentially more important as finding more optimal solutions earlier allows to prune more nodes from the exponentially expanding search tree.
Specifically we evaluate our heuristic at every node for the first 250 selections, then at every tenth node for the next 750 nodes, and finally switch to classical selectors for the remaining nodes.\footnote{This accounts for the ``phase-transition'' in MIP solvers where optimality needs to be proved by closing the remaining branches although the theoretically optimal point is already found~\citep{MORRISON201679}. Note that with a tuned implementation we could run our method for more nodes, where we expect further improvements.}

\subsection{Data Generation \& Agent Training}
\label{subsec:MethodData}

In training MIPs, a critical challenge lies in generating sufficiently complex training problems. 
First, to learn from interesting structures, we need to decide on some specific problem, whose \eg satisfiability is knowable as generating random constraint matrices will likely generate empty polyhedrons, or polyhedrons with many eliminable constraints (e.g., in the constraint set consisting of $c^Tx\leq b$ and $c^Tx\leq b+\rho$ with $\rho\neq 0$ one constraint is always eliminable).
This may seem unlikely, but notice how we can construct aligned $c$ vectors by linearly combining different rows (just like in LP-dual formulations).
In practice, selecting a sufficiently large class of problems may be enough as during the branch-and-cut process many sub-problems of different characteristics are generated.
Since our algorithm naturally decomposes the problem into sub-trees, we can assume any policy that performs well on the entire tree also performs well on sub-polyhedra generated during the branch-and-cut.

For this reason we consider the large class of Traveling Salesman Problem (TSP), which itself has rich use-cases in planning and logistics, but also in optimal control, the manufacturing of microchips and DNA sequencing (see~\citet{cook2011traveling}).
The TSP problem consists of finding a round-trip path in a weighted graph, such that every vertex is visited exactly once, and the total path-length is minimal (for more details and a mathematical formulation, see Appendix~\ref{sec:tsp-as-milp}).

For training, we would like to use random instances of TSP but generating them can be challenging.
Random sampling of distance matrices often results in easy problem instances, which do not challenge the solver. 
Consequently, significant effort has been devoted to devising methods for generating random but hard instances, particularly for problems like the TSP, where specific generators for challenging problems have been designed (see~\citet{vercesiGenerationMetricTSP2023} and~\citet{randomCut}).
However, for our specific use cases, these provably hard problems may not be very informative as they rarely contain efficiently selectable nodes. 

To generate these intermediary-difficult problems, we adopt a multi-step approach: We begin by generating random instances and then apply some mutation techniques~\citep{10.1145/3299904.3340307} to introduce variations, and ensure diversity within the problem set.
Next, we select the instance of median-optimality gap from the pool, which produces an instance of typical difficulty.
The optimality gap, representing the best normalized difference between the lower and upper bound for a solution found during the solver's budget-restricted execution, serves as a crucial metric to assess difficulty. 
This method is used to produce 200 intermediary-difficult training instances.

To ensure the quality of candidate problems, we exclude problems with more than $100\%$ or zero optimality gap, as these scenarios present challenges in reward assignment during RL.
To reduce overall variance of our training, we limit the ground-truth variance in optimality gap.
Additionally, we impose a constraint on the minimum number of nodes in the problems, discarding every instance with less than 100 nodes. 
This is essential as we do not expect such small problems to give clean reward signals to the reinforcement learner.

\section{Experiments}\label{sec:Experiments}

For our experiments we consider the instances of \textsc{TSPLIB}~\citep{TSPLib} and  \textsc{MIPLIB}~\citep{MIPLIB} which are one of the most used datasets for benchmarking MIP frameworks and thusly form a strong baseline to test against.
We further test against the UFLP instance generator by \citep{Kochetov2005}, which specifically produces instances hard to solve for branch-and-bound, and against \textsc{MINLPLIB}~\citep{MINLPLIB}, which contains \emph{mixed integer nonlinear programs}, to show generalization towards very foreign problems.
\iftrue\footnote{Source code: \url{https://github.com/MattAlexMiracle/BranchAndBoundBisim}}\else \fi
All training and experiments were done fully on a ryzen 7 5800x3d CPU and completes training in about 6\si{h}.
We were generally not limited by compute, but rather by the fact that every policy rollout takes, by definition, 45\si{s} regardless of hardware used.


\subsection{Baselines}
\label{subsec:Experiments:Baselines}

We run both our method and SCIP for 45s.
We then filter out all runs where SCIP has managed to explore less than 5 nodes, as in these runs even perfect node selection makes no difference in performance.
If we included those in our average, we would have a significant number of lines where our node-selector has zero advantage over the traditional SCIP one, not because our selector is better or worse than SCIP, but simply because it wasn't called in the first place. 
We set this time-limit relatively low as our prototype selector only runs at the beginning of the solver process, meaning that over time the effects of the traditional solver take over. 

{Additionally, we also consider an evaluation with a 5\si{min} time limit. For those runs, the limit until we switch to the naive node selector is set to 650 nodes to account for the higher time limit.
In general, the relative performance of our method compared to the baseline increases with the 5\si{min} budget.}


{
Finally, we also benchmark against the previous state-of-the-art by \citet{labassiLearningCompareNodes}, which represents optimal node selection as a comparison problem where a siamese network is used to compare different leaf nodes against each other and picking the ``highest'' node as the next selection.
}

\subsection{Results}

While all results can be found in Appendix~\ref{sec:fullResults} we report an aggregated view for each benchmark in Table~\ref{tab:otherTests}.
In addition to our ``reward'' metric we report the winning ratio of our method over the baseline, and the geometric mean of the gaps at the end of solving (lower is better).

\begin{table*}[t]
    \small
    \centering
    \caption{Performance across benchmarks (the policy only saw TSP instances during training).
    }
    \vspace{-2mm}
    \resizebox{\textwidth}{!}{
    \begin{tabular}{lcccccc}\toprule
         Benchmark                &Reward&  Win-rate & geo-mean Ours & geo-mean SCIP \\\midrule
         TSPLIB~\citep{TSPLib}     &0.184            & 0.50                     & 0.931                          & 0.957 \\
         UFLP~\citep{Kochetov2005} &0.078          & 0.636                    & 0.491                          & 0.520 \\
         MINLPLib~\citep{MINLPLIB} &0.487             & 0.852                    & 28.783                         & 31.185\\
         MIPLIB~\citep{MIPLIB}     &0.140        & 0.769                    & 545.879                        & 848.628\\
      TSPLIB@5min &0.192            & 0.600                    & 1.615                          & 2.000 \\
     MINLPlib@5min&0.486            & 0.840                    & 17.409                         & 20.460\\
     MINLPlib@5min&0.150            & 0.671                    & 66.861                         & 106.400\\
         \bottomrule
    \end{tabular}
    }
    \label{tab:otherTests}
\end{table*}

For benchmarking and training, we leave all settings, such as presolvers, primal heuristics, diving heuristics, constraint specialisations, etc. at their default settings to allow the baseline to perform best.
All 
instances are solved using the same model without any fine-tuning.
We expect that tuning, e.g., the aggressiveness of primal heuristics, increases the performance of our method, as it decreases the relative cost of evaluating a neural network, but for the sake of comparison we use the default parameters for all our tests. 
We train our node selection policy on problem instances according to Sec.~\ref{subsec:MethodData} and apply it on problems from different benchmarks.

First, we will discuss TSPLIB itself, which while dramatically more complex than our selected training instances, still contains instances from the same problem family as the training set (Sec.~\ref{sec:exp:TSPLIB}).
Second, we consider instances of the Uncapacitated Facility Location Problem (UFLP) as generated by \cite{Kochetov2005}'s problem generator.
These problems are designed to be particulary challenging to branch-and-bound solvers due to their large optimality gap (Sec.~\ref{sec:exp:UFLP}).
While the first two benchmarks focused on specific problems (giving you a notion of how well the algorithm does on the problem itself) we next consider ```meta-benchmarks'' that consist of many different problems, but relatively few instances of each. MINLPLIB~\citep{MINLPLIB} is a meta-benchmark for \emph{nonlinear} mixed-integer programming (Sec.~\ref{sec:exp:MINLPLIB}), and MIPLIB~\citep{MIPLIB} a benchmark for mixed integer programming (Sec.~\ref{sec:exp:MIPLIB}).
We also consider generalisation against the uncapacitated facility location problem using a strong instance generator, see Appendix~\ref{sec:exp:UFLP}.
Our benchmarks are diverse and complex and allow to compare algorithmic improvements in state-of-the-art solvers.

\subsubsection{TSPLIB}
\label{sec:exp:TSPLIB}


{
From an aggregative viewpoint we outperform the SCIP node selection by $\approx 20\%$ in reward. 
However, the overall ``win-rate'' is only $50\%$ as the mean-reward is dominated by instances where our solver has significant performance improvements: When our method looses, it looses by relatively little (\eg \texttt{att48}: ours $0.287$ vs base $0.286$), while when it wins, it usually does so by a larger margin.
}

Qualitatively, it is particularly interesting to study the problems our method still looses significantly against SCIP (in four cases).
A possible reason why our method significantly underperforms on \texttt{Dantzig42} is that our implementation is just too slow, considering that the baseline manages to evaluate $\approx 40\%$ more nodes. A similar observation can be made on \texttt{eil51} where the baseline manages to complete $5\times$  more nodes.
\texttt{rd100} is also similar to \texttt{Dantzig} and \texttt{eil51} as the baseline is able to explore $60\%$ more nodes.
\texttt{KroE100} is the first instance our method looses against SCIP, despite exploring a roughly equal amount of nodes. We believe that this is because our method commits to the wrong subtree early and never manages to correct into the proper subtree.
Ignoring these four failure cases, our method is either on par (up to stochasticity of the algorithm) or exceeds the baseline significantly.

It is also worthwhile to study the cases where both the baseline and our method hit $0$ optimality gap.
{Both algorithms reaching zero optimality gap can be seen as somewhat of a special case, since soley looking at the solution value is insufficient to figure out which method performs better in practice.}
A quick glance at instances like \texttt{bayg29}, \texttt{fri26}, \texttt{swiss42} or \texttt{ulysses16} shows that our method tends to finish these problems with significantly fewer nodes explored.
This is not captured by any of our metrics since those only look at solution quality, not the efficiency of reaching that solution.
If the quality of the baseline and ours is the same, it makes sense to look at solution efficiency instead.
Qualitatively, instances like \texttt{bayg29} manage to reach the optimum in only $\frac{1}{3}$ the number of explored nodes, which showcases a significant improvement in node-selection quality.
It is worth noting that, due to the different optimization costs for different nodes, it not always true that evaluating fewer nodes is faster in wall-clock time.
In practice, ``fewer nodes is better'' seems to be a good rule-of-thumb to check algorithmic efficiency.

{
\subsection{UFLP}
The UFLP benchmark designed by \citet{Kochetov2005} is specifically built to be hard to solve by branch-and-bound solvers due to its large duality gap.
Despite this, our method manages to outperform the baseline significantly.
This is meaningful since this benchmark is a specially designed worst-case scenario: The fact that our method still outperforms the baseline provides good evidence of the efficacy of the method.
}
\subsubsection{MINLPLIB}
\label{sec:exp:MINLPLIB}

We now consider MINLPs. To solve these, SCIP and other solvers use branching techniques that cut nonlinear (often convex) problems from a relaxed master-relaxation towards true solutions.
We consider MINLPLib~\citep{MINLPLIB}, a meta-benchmark consisting of hundreds of synthetic and real-world MINLP instances of varying different types and sizes.
{As some instances take hours to solve even a single node, we filter out all problems with fewer than 5 nodes, as the performance in those cases is independent of the node-selectors performance}
(Full results Appendix~\ref{sec:MINLP}).

Our method still manages to outperform SCIP, even on MINLPs, despite never having seen a single MINLP problem before, see Table~\ref{tab:otherTests}.
Qualitatively, our method either outperforms or is on par with the vast majority of problems, but also loses significantly in some problems, greatly decreasing the average.
Studying the cases our method looses convincingly (see app.~\ref{sec:MINLP}), we find a similar case as in TSPLIB, where the baseline implementation is so much more optimized that significantly more nodes can get explored.
{
We suspect the reason our method has the biggest relative improvements on MINLPs is due to the fact that existing \eg pseudocost based selection methods do not perform as well on spatial-branch-and-bound tasks.
}

We expect features specifically tuned for nonlinear problems to increase performance by additional percentage points, but as feature selection is orthogonal to the actual algorithm design, we leave more thorough discussion of this to future work~\footnote{We are not aware of a learned BnB node-selection heuristic used for MINLPs, so guidance towards feature selection doesn't exist yet. Taking advantage of them presumably also requires to train on nonlinear problems.}.

\subsubsection{MIPLIB}
\label{sec:exp:MIPLIB}

Last, but not least we consider the meta-benchmark MIPLIB~\citep{MIPLIB}, which consists of hundreds of real-world mixed-integer programming problems of varying size, complexity, and hardness. Our method is either close to or exceeds the performance of SCIP, see Table~\ref{tab:otherTests}.

Considering per-instance results, we see similar patterns as in previous failure cases:
Often we underperform on instances that need to close many nodes, as our method's throughput lacks behind that of SCIP.
We expect that a more efficient implementation alleviates the issues in those cases.

We also see challenges in problems that are far from the training distribution, specifically satisfaction problems.
Consider \texttt{fhnw-binpack4-48}, were the baseline yields an optimality gap of $0$ while we end at $+\infty$.
This is due to the design of the problem: Instead of a classical optimization problem, this is a satisfaction problem, where not an optimal value, but \emph{any} feasible value is searched, i.e., we either yield a gap of $0$ (solution found), or a gap of $+\infty$ (solution not found), as no other gap is possible.
Notably, these kinds of problems may pose a challenge for our algorithm, as the node-pruning dynamics of satisfying MIPs are different than the one for optimizing MIPs: Satisfying MIPs can only rarely prune nodes since, by definition, no intermediary primally valid solutions are ever found.

{
\subsection{Comparison against ``Learning to compare nodes''}\label{sec:labassi}

\begin{table*}[!htb]
    \centering
    \caption{Comparison against \citep{labassiLearningCompareNodes}. Note that we use a \emph{single} network, evaluated out-of-distribution, while \citep{labassiLearningCompareNodes} uses \emph{different} networks, all trained on that specific type of mixed integer linear program.
    For all metrics, lower is better.
    }
    \resizebox{\textwidth}{!}{
    \begin{tabular}{l cc cc cc cc}\toprule
                 &\multicolumn{2}{c}{FMCNF} & \multicolumn{2}{c}{GISP} & \multicolumn{2}{c}{WPMS}\\\midrule
                 &Nodes              & Runtime           & Nodes & Runtime          & Nodes & Runtime         \\\midrule
    Labassi FMCNF& 339.53 $\pm$ 5.90 & 29.45 $\pm$ 2.13  & ---    & ---               & ---    & ---              \\
    Labassi GISP & --- & ---  & 1219 $\pm$ 1.73    & 26.50 $\pm$ 1.55               & ---    & ---              \\
    Labassi WPMS & --- & ---   & ---    & ---               & \textbf{215.26 $\pm$ 1.97}    & 10.46 $\pm$ 1.56              \\
    Ours         & \textbf{187.33 $\pm$ 3.67} & \textbf{19.99 $\pm$ 2.25}   &  \textbf{1216.16 $\pm$ 1.91}   & \textbf{16.94 $\pm$ 1.49}               & 221.73 $\pm$ 1.78    & \textbf{8.04 $\pm$ 1.72}              \\\bottomrule
    \end{tabular}}

    \label{tab:labassi}
\end{table*}

Aside from comparisons against SCIP, we also compare against the previous state-of-the-art method by \citet{labassiLearningCompareNodes}. One complication when benchmarking against \citet{labassiLearningCompareNodes} was that labassi assumes a \emph{seperate model for each problem type}, while our method is already flexible enough to handle \emph{all problem types within a single model}.
Since \citet{labassiLearningCompareNodes} needs to re-train for each instance type and only works for linear problems, we cannot test it against MIPLIB and MINLPLIB.
Instead, we reproduce all benchmarks used by labassi (each with their own separate models), and compare against them against our single model. 

As one can see in Table~\ref{tab:labassi}, we convincingly beat the prior work on every single benchmark (aside of WPMS node count), despite our method never having seen any of these problem types during training. 
This is surprising because \citep{labassiLearningCompareNodes} should have a convincing advantage due to the fact that not only did they train a dedicated agents for every single one of their problems, they also use the same generator for their training and testing instances.
This implies that your single solver, evaluated out-of-distribution, manages to outperform the specialized agents proposed by \citep{labassiLearningCompareNodes} evaluated in-distribution.
In general, our method improves upon \citep{labassiLearningCompareNodes} by between $30\%$ and $56\%$ wrt. runtime. 
}
\section{Limitations}
There are still many open questions.
Feature selection remains an area where we expect significant improvements, especially for nonlinear programming, which contemporary methods do not account for.
We also expect significant improvements in performance through code optimization.
An important area for research lies in generalized instance generation: Instead of focusing on single domain instances for training (e.g. from TSP), an instance generator should create problem instances with consistent, but varying levels of difficulty across different problem domains. 
Further, the number of nodes used before switching to classical node selectors is only done heuristically. Finding optimal switching points between different node selectors is still an open problem even beyond learned solutions and represents an interesting place for further research.

\section{Conclusion}\label{sec:Conclusion}

We have proposed a novel approach to branch-and-bound node selection, that uses the structure of the branch-and-bound process and reinforcement learning to convincingly beat classical SCIP and learnt node selectors.
By aligning our model with the branch-and-bound tree structure, we have demonstrated the potential to develop a versatile heuristic that can be applied across various optimization problem domains, despite being trained on a narrow set of instances.
To our knowledge, this is the first demonstration of learned node selection to mixed-integer (nonlinear) programming.

\subsubsection*{Acknowledgments}
This work was supported by the Bavarian Ministry for Economic Affairs, Infrastructure, Transport and Technology through the Center for Analytics-Data-Applications (ADA-Center) within the framework of ``BAYERN DIGITAL II''.


\bibliography{neurips2024_conference}

\begin{thebibliography}{40}
\providecommand{\natexlab}[1]{#1}
\providecommand{\url}[1]{\texttt{#1}}
\expandafter\ifx\csname urlstyle\endcsname\relax
  \providecommand{\doi}[1]{doi: #1}\else
  \providecommand{\doi}{doi: \begingroup \urlstyle{rm}\Url}\fi

\bibitem[Andrychowicz et~al.(2020)Andrychowicz, Raichuk, Sta{\'n}czyk, Orsini,
  Girgin, Marinier, Hussenot, Geist, Pietquin, Michalski, Gelly, and
  Bachem]{andrychowiczWhatMattersOnPolicy2020}
Andrychowicz, M., Raichuk, A., Sta{\'n}czyk, P., Orsini, M., Girgin, S.,
  Marinier, R., Hussenot, L., Geist, M., Pietquin, O., Michalski, M., Gelly,
  S., and Bachem, O.
\newblock What {{Matters In On-Policy Reinforcement Learning}}? {{A Large-Scale
  Empirical Study}}, June 2020.

\bibitem[Ba et~al.(2016)Ba, Kiros, and Hinton]{ba2016layer}
Ba, J.~L., Kiros, J.~R., and Hinton, G.~E.
\newblock Layer normalization.
\newblock \emph{arXiv preprint arXiv:1607.06450}, 2016.

\bibitem[Bachlechner et~al.(2020)Bachlechner, Majumder, Mao, Cottrell, and
  McAuley]{Bachlechner2020ReZeroIA}
Bachlechner, T.~C., Majumder, B.~P., Mao, H.~H., Cottrell, G., and McAuley, J.
\newblock Rezero is all you need: Fast convergence at large depth.
\newblock In \emph{Conference on Uncertainty in Artificial Intelligence}, 2020.
\newblock URL \url{https://api.semanticscholar.org/CorpusID:212644626}.

\bibitem[Bayliss et~al.(2017)Bayliss, Maere, Atkin, and
  Paelinck]{Bayliss2017ASS}
Bayliss, C., Maere, G.~D., Atkin, J. A.~D., and Paelinck, M.
\newblock A simulation scenario based mixed integer programming approach to
  airline reserve crew scheduling under uncertainty.
\newblock \emph{Annals of Operations Research}, 252:\penalty0 335--363, 2017.

\bibitem[Bestuzheva et~al.(2021{\natexlab{a}})Bestuzheva, Besan{\c{c}}on, Chen,
  Chmiela, Donkiewicz, van Doornmalen, Eifler, Gaul, Gamrath, Gleixner,
  Gottwald, Graczyk, Halbig, Hoen, Hojny, van~der Hulst, Koch, L{\"u}bbecke,
  Maher, Matter, M{\"u}hmer, M{\"u}ller, Pfetsch, Rehfeldt, Schlein,
  Schl{\"o}sser, Serrano, Shinano, Sofranac, Turner, Vigerske, Wegscheider,
  Wellner, Weninger, and Witzig]{BestuzhevaEtal2021OO}
Bestuzheva, K., Besan{\c{c}}on, M., Chen, W.-K., Chmiela, A., Donkiewicz, T.,
  van Doornmalen, J., Eifler, L., Gaul, O., Gamrath, G., Gleixner, A.,
  Gottwald, L., Graczyk, C., Halbig, K., Hoen, A., Hojny, C., van~der Hulst,
  R., Koch, T., L{\"u}bbecke, M., Maher, S.~J., Matter, F., M{\"u}hmer, E.,
  M{\"u}ller, B., Pfetsch, M.~E., Rehfeldt, D., Schlein, S., Schl{\"o}sser, F.,
  Serrano, F., Shinano, Y., Sofranac, B., Turner, M., Vigerske, S.,
  Wegscheider, F., Wellner, P., Weninger, D., and Witzig, J.
\newblock {The SCIP Optimization Suite 8.0}.
\newblock Technical report, Optimization Online, December 2021{\natexlab{a}}.

\bibitem[Bestuzheva et~al.(2021{\natexlab{b}})Bestuzheva, Besan{\c{c}}on, Chen,
  Chmiela, Donkiewicz, van Doornmalen, Eifler, Gaul, Gamrath, Gleixner,
  Gottwald, Graczyk, Halbig, Hoen, Hojny, van~der Hulst, Koch, L{\"u}bbecke,
  Maher, Matter, M{\"u}hmer, M{\"u}ller, Pfetsch, Rehfeldt, Schlein,
  Schl{\"o}sser, Serrano, Shinano, Sofranac, Turner, Vigerske, Wegscheider,
  Wellner, Weninger, and Witzig]{BestuzhevaEtal2021ZR}
Bestuzheva, K., Besan{\c{c}}on, M., Chen, W.-K., Chmiela, A., Donkiewicz, T.,
  van Doornmalen, J., Eifler, L., Gaul, O., Gamrath, G., Gleixner, A.,
  Gottwald, L., Graczyk, C., Halbig, K., Hoen, A., Hojny, C., van~der Hulst,
  R., Koch, T., L{\"u}bbecke, M., Maher, S.~J., Matter, F., M{\"u}hmer, E.,
  M{\"u}ller, B., Pfetsch, M.~E., Rehfeldt, D., Schlein, S., Schl{\"o}sser, F.,
  Serrano, F., Shinano, Y., Sofranac, B., Turner, M., Vigerske, S.,
  Wegscheider, F., Wellner, P., Weninger, D., and Witzig, J.
\newblock {The SCIP Optimization Suite 8.0}.
\newblock ZIB-Report 21-41, Zuse Institute Berlin, December 2021{\natexlab{b}}.

\bibitem[Bestuzheva et~al.(2023)Bestuzheva, Chmiela, M{\"u}ller, Serrano,
  Vigerske, and Wegscheider]{bestuzhevaGlobalOptimizationMixedInteger2023}
Bestuzheva, K., Chmiela, A., M{\"u}ller, B., Serrano, F., Vigerske, S., and
  Wegscheider, F.
\newblock Global {{Optimization}} of {{Mixed-Integer Nonlinear Programs}} with
  {{SCIP}} 8, January 2023.

\bibitem[Bossek et~al.(2019)Bossek, Kerschke, Neumann, Wagner, Neumann, and
  Trautmann]{10.1145/3299904.3340307}
Bossek, J., Kerschke, P., Neumann, A., Wagner, M., Neumann, F., and Trautmann,
  H.
\newblock Evolving diverse tsp instances by means of novel and creative
  mutation operators.
\newblock In \emph{Proceedings of the 15th ACM/SIGEVO Conference on Foundations
  of Genetic Algorithms}, FOGA '19, pp.\  58–71, New York, NY, USA, 2019.
  Association for Computing Machinery.
\newblock ISBN 9781450362542.
\newblock \doi{10.1145/3299904.3340307}.

\bibitem[Bussieck et~al.(2003)Bussieck, Drud, and Meeraus]{MINLPLIB}
Bussieck, M.~R., Drud, A.~S., and Meeraus, A.
\newblock {MINLPLib}{\textemdash}a collection of test models for mixed-integer
  nonlinear programming.
\newblock \emph{{INFORMS} Journal on Computing}, 15\penalty0 (1):\penalty0
  114--119, February 2003.
\newblock \doi{10.1287/ijoc.15.1.114.15159}.

\bibitem[Cook et~al.(2011)Cook, Applegate, Bixby, and
  Chvatal]{cook2011traveling}
Cook, W.~J., Applegate, D.~L., Bixby, R.~E., and Chvatal, V.
\newblock \emph{The traveling salesman problem: a computational study}.
\newblock Princeton university press, 2011.

\bibitem[Dantzig(1982)]{DANTZIG198243}
Dantzig, G.~B.
\newblock Reminiscences about the origins of linear programming.
\newblock \emph{Operations Research Letters}, 1\penalty0 (2):\penalty0 43--48,
  1982.
\newblock ISSN 0167-6377.
\newblock \doi{https://doi.org/10.1016/0167-6377(82)90043-8}.

\bibitem[Dayal et~al.(2022)Dayal, Cenkeramaddi, and Jha]{DAYAL2022109241}
Dayal, A., Cenkeramaddi, L.~R., and Jha, A.
\newblock Reward criteria impact on the performance of reinforcement learning
  agent for autonomous navigation.
\newblock \emph{Applied Soft Computing}, 126:\penalty0 109241, 2022.
\newblock ISSN 1568-4946.
\newblock \doi{https://doi.org/10.1016/j.asoc.2022.109241}.
\newblock URL
  \url{https://www.sciencedirect.com/science/article/pii/S1568494622004586}.

\bibitem[Etheve et~al.(2020)Etheve, Al{\`e}s, Bissuel, Juan, and
  Kedad-Sidhoum]{Etheve2020ReinforcementLF}
Etheve, M., Al{\`e}s, Z., Bissuel, C., Juan, O., and Kedad-Sidhoum, S.
\newblock Reinforcement learning for variable selection in a branch and bound
  algorithm.
\newblock \emph{ArXiv}, abs/2005.10026, 2020.
\newblock URL \url{https://api.semanticscholar.org/CorpusID:211551730}.

\bibitem[Floudas \& Lin(2005)Floudas and Lin]{Floudas2005MixedIL}
Floudas, C.~A. and Lin, X.
\newblock Mixed integer linear programming in process scheduling: Modeling,
  algorithms, and applications.
\newblock \emph{Annals of Operations Research}, 139:\penalty0 131--162, 2005.

\bibitem[Gasse et~al.(2022)Gasse, Cappart, Charfreitag, Charlin, Ch'etelat,
  Chmiela, Dumouchelle, Gleixner, Kazachkov, Khalil, Lichocki, Lodi, Lubin,
  Maddison, Morris, Papageorgiou, Parjadis, Pokutta, Prouvost, Scavuzzo,
  Zarpellon, Yangm, Lai, Wang, Luo, Zhou, Huang, Shao, Zhu, Zhang, Quan, Cao,
  Xu, Huang, Zhou, Binbin, Minggui, Hao, Zhiyu, Zhiwu, and Kun]{Gasse2022TheML}
Gasse, M., Cappart, Q., Charfreitag, J., Charlin, L., Ch'etelat, D., Chmiela,
  A., Dumouchelle, J., Gleixner, A.~M., Kazachkov, A.~M., Khalil, E.~B.,
  Lichocki, P., Lodi, A., Lubin, M., Maddison, C.~J., Morris, C., Papageorgiou,
  D.~J., Parjadis, A., Pokutta, S., Prouvost, A., Scavuzzo, L., Zarpellon, G.,
  Yangm, L., Lai, S., Wang, A., Luo, X., Zhou, X., Huang, H., Shao, S.~C., Zhu,
  Y., Zhang, D., Quan, T.~M., Cao, Z., Xu, Y., Huang, Z., Zhou, S., Binbin, C.,
  Minggui, H., Hao, H., Zhiyu, Z., Zhiwu, A., and Kun, M.
\newblock The machine learning for combinatorial optimization competition
  (ml4co): Results and insights.
\newblock \emph{ArXiv}, abs/2203.02433, 2022.
\newblock URL \url{https://api.semanticscholar.org/CorpusID:247245014}.

\bibitem[Gelfand \& Smith(1990)Gelfand and
  Smith]{f47e9344-75dc-3782-a811-9e51a952e8bf}
Gelfand, A.~E. and Smith, A. F.~M.
\newblock Sampling-based approaches to calculating marginal densities.
\newblock \emph{Journal of the American Statistical Association}, 85\penalty0
  (410):\penalty0 398--409, 1990.
\newblock ISSN 01621459.
\newblock URL \url{http://www.jstor.org/stable/2289776}.

\bibitem[Gleixner et~al.(2021)Gleixner, Hendel, Gamrath, Achterberg, Bastubbe,
  Berthold, Christophel, Jarck, Koch, Linderoth, L{\"u}bbecke, Mittelmann,
  Ozyurt, Ralphs, Salvagnin, and Shinano]{MIPLIB}
Gleixner, A., Hendel, G., Gamrath, G., Achterberg, T., Bastubbe, M., Berthold,
  T., Christophel, P., Jarck, K., Koch, T., Linderoth, J., L{\"u}bbecke, M.,
  Mittelmann, H.~D., Ozyurt, D., Ralphs, T.~K., Salvagnin, D., and Shinano, Y.
\newblock {{MIPLIB}} 2017: Data-driven compilation of the 6th~mixed-integer
  programming library.
\newblock \emph{Mathematical Programming Computation}, 13\penalty0
  (3):\penalty0 443--490, September 2021.
\newblock ISSN 1867-2957.
\newblock \doi{10.1007/s12532-020-00194-3}.

\bibitem[He et~al.(2014)He, Daume~III, and Eisner]{heLearningSearchBranch}
He, H., Daume~III, H., and Eisner, J.~M.
\newblock Learning to search in branch and bound algorithms.
\newblock In Ghahramani, Z., Welling, M., Cortes, C., Lawrence, N., and
  Weinberger, K. (eds.), \emph{Advances in Neural Information Processing
  Systems}, volume~27. Curran Associates, Inc., 2014.

\bibitem[Kingma \& Welling(2013)Kingma and Welling]{Kingma2013AutoEncodingVB}
Kingma, D.~P. and Welling, M.
\newblock Auto-encoding variational bayes.
\newblock \emph{CoRR}, abs/1312.6114, 2013.
\newblock URL \url{https://api.semanticscholar.org/CorpusID:216078090}.

\bibitem[Kochetov \& Ivanenko(2005)Kochetov and Ivanenko]{Kochetov2005}
Kochetov, Y. and Ivanenko, D.
\newblock \emph{Computationally Difficult Instances for the Uncapacitated
  Facility Location Problem}, pp.\  351--367.
\newblock Springer US, Boston, MA, 2005.
\newblock ISBN 978-0-387-25383-1.
\newblock \doi{10.1007/0-387-25383-1_16}.

\bibitem[Labassi et~al.(2022)Labassi, Chételat, and
  Lodi]{labassiLearningCompareNodes}
Labassi, A.~G., Chételat, D., and Lodi, A.
\newblock Learning to compare nodes in branch and bound with graph neural
  networks.
\newblock In \emph{Advances in Neural Information Processing Systems 35}, 2022.

\bibitem[Loshchilov \& Hutter(2017)Loshchilov and
  Hutter]{Loshchilov2017FixingWD}
Loshchilov, I. and Hutter, F.
\newblock Fixing weight decay regularization in adam.
\newblock \emph{ArXiv}, abs/1711.05101, 2017.
\newblock URL \url{https://api.semanticscholar.org/CorpusID:3312944}.

\bibitem[Menon et~al.(2013)Menon, Nabil, and Narasimhan]{MENON2013690}
Menon, G., Nabil, M., and Narasimhan, S.
\newblock Branch and bound algorithm for optimal sensor network design.
\newblock \emph{IFAC Proceedings Volumes}, 46\penalty0 (32):\penalty0 690--695,
  2013.
\newblock ISSN 1474-6670.
\newblock \doi{https://doi.org/10.3182/20131218-3-IN-2045.00143}.
\newblock 10th IFAC International Symposium on Dynamics and Control of Process
  Systems.

\bibitem[Miller et~al.(1960)Miller, Tucker, and Zemlin]{MTZ}
Miller, C.~E., Tucker, A.~W., and Zemlin, R.~A.
\newblock Integer programming formulation of traveling salesman problems.
\newblock \emph{J. ACM}, 7\penalty0 (4):\penalty0 326–329, oct 1960.
\newblock ISSN 0004-5411.
\newblock \doi{10.1145/321043.321046}.

\bibitem[Mittelmann(2021)]{DecisonTreeOptimization}
Mittelmann, H.~D.
\newblock Decison {{Tree}} for {{Optimization Software}}.
\newblock \url{https://plato.asu.edu/sub/nlores.html}, 2021.
\newblock Accessed: 2023-09-02.

\bibitem[Mnih et~al.(2015)Mnih, Kavukcuoglu, Silver, Rusu, Veness, Bellemare,
  Graves, Riedmiller, Fidjeland, Ostrovski, Petersen, Beattie, Sadik,
  Antonoglou, King, Kumaran, Wierstra, Legg, and Hassabis]{Mnih2015}
Mnih, V., Kavukcuoglu, K., Silver, D., Rusu, A.~A., Veness, J., Bellemare,
  M.~G., Graves, A., Riedmiller, M., Fidjeland, A.~K., Ostrovski, G., Petersen,
  S., Beattie, C., Sadik, A., Antonoglou, I., King, H., Kumaran, D., Wierstra,
  D., Legg, S., and Hassabis, D.
\newblock Human-level control through deep reinforcement learning.
\newblock \emph{Nature}, 518\penalty0 (7540):\penalty0 529--533, February 2015.
\newblock \doi{10.1038/nature14236}.
\newblock URL \url{https://doi.org/10.1038/nature14236}.

\bibitem[Mnih et~al.(2016)Mnih, Badia, Mirza, Graves, Lillicrap, Harley,
  Silver, and Kavukcuoglu]{Mnih2016AsynchronousMF}
Mnih, V., Badia, A.~P., Mirza, M., Graves, A., Lillicrap, T.~P., Harley, T.,
  Silver, D., and Kavukcuoglu, K.
\newblock Asynchronous methods for deep reinforcement learning.
\newblock In \emph{International Conference on Machine Learning}, 2016.
\newblock URL \url{https://api.semanticscholar.org/CorpusID:6875312}.

\bibitem[Morrison et~al.(2016)Morrison, Jacobson, Sauppe, and
  Sewell]{MORRISON201679}
Morrison, D.~R., Jacobson, S.~H., Sauppe, J.~J., and Sewell, E.~C.
\newblock Branch-and-bound algorithms: A survey of recent advances in
  searching, branching, and pruning.
\newblock \emph{Discrete Optimization}, 19:\penalty0 79--102, 2016.
\newblock ISSN 1572-5286.
\newblock \doi{https://doi.org/10.1016/j.disopt.2016.01.005}.

\bibitem[Otten \& Dechter(2012)Otten and Dechter]{ottenCaseStudyComplexity}
Otten, L. and Dechter, R.
\newblock A {{Case Study}} in {{Complexity Estimation}}: {{Towards Parallel
  Branch-and-Bound}} over {{Graphical Models}}.
\newblock \emph{Uncertainty in Artificial Intelligence - Proceedings of the
  28th Conference, UAI 2012}, October 2012.

\bibitem[Parsonson et~al.(2022)Parsonson, Laterre, and
  Barrett]{Parsonson2022ReinforcementLF}
Parsonson, C. W.~F., Laterre, A., and Barrett, T.~D.
\newblock Reinforcement learning for branch-and-bound optimisation using
  retrospective trajectories.
\newblock In \emph{AAAI Conference on Artificial Intelligence}, 2022.
\newblock URL \url{https://api.semanticscholar.org/CorpusID:249192146}.

\bibitem[Rardin et~al.(1993)Rardin, Tovey, and Pilcher]{randomCut}
Rardin, R.~L., Tovey, C.~A., and Pilcher, M.~G.
\newblock Analysis of a {{Random Cut Test Instance Generator}} for the {{TSP}}.
\newblock In \emph{Complexity in {{Numerical Optimization}}}, pp.\  387--405.
  {WORLD SCIENTIFIC}, July 1993.
\newblock ISBN 978-981-02-1415-9.
\newblock \doi{10.1142/9789814354363_0017}.

\bibitem[Reinelt(1991)]{TSPLib}
Reinelt, G.
\newblock {TSPLIB}{\textemdash}a traveling salesman problem library.
\newblock \emph{{ORSA} Journal on Computing}, 3\penalty0 (4):\penalty0
  376--384, November 1991.
\newblock \doi{10.1287/ijoc.3.4.376}.

\bibitem[Robert \& Roberts(2021)Robert and
  Roberts]{Robert2021RaoBlackwellizationIT}
Robert, C.~P. and Roberts, G.~O.
\newblock Rao-blackwellization in the mcmc era.
\newblock 2021.
\newblock URL \url{https://api.semanticscholar.org/CorpusID:230437867}.

\bibitem[Schulman et~al.(2015)Schulman, Moritz, Levine, Jordan, and
  Abbeel]{schulmanHighDimensionalContinuousControl2015}
Schulman, J., Moritz, P., Levine, S., Jordan, M.~I., and Abbeel, P.
\newblock High-{{Dimensional Continuous Control Using Generalized Advantage
  Estimation}}.
\newblock \emph{CoRR}, June 2015.

\bibitem[Schulman et~al.(2017)Schulman, Wolski, Dhariwal, Radford, and
  Klimov]{Schulman2017ProximalPO}
Schulman, J., Wolski, F., Dhariwal, P., Radford, A., and Klimov, O.
\newblock Proximal policy optimization algorithms.
\newblock \emph{ArXiv}, abs/1707.06347, 2017.

\bibitem[Silver et~al.(2017)Silver, Hubert, Schrittwieser, Antonoglou, Lai,
  Guez, Lanctot, Sifre, Kumaran, Graepel, Lillicrap, Simonyan, and
  Hassabis]{Silver2017MasteringCA}
Silver, D., Hubert, T., Schrittwieser, J., Antonoglou, I., Lai, M., Guez, A.,
  Lanctot, M., Sifre, L., Kumaran, D., Graepel, T., Lillicrap, T.~P., Simonyan,
  K., and Hassabis, D.
\newblock Mastering chess and shogi by self-play with a general reinforcement
  learning algorithm.
\newblock \emph{ArXiv}, abs/1712.01815, 2017.
\newblock URL \url{https://api.semanticscholar.org/CorpusID:33081038}.

\bibitem[Sutton \& Barto(2018)Sutton and Barto]{sutton2018reinforcement}
Sutton, R.~S. and Barto, A.~G.
\newblock \emph{Reinforcement learning: An introduction}.
\newblock MIT press, 2018.

\bibitem[Vercesi et~al.(2023)Vercesi, Gualandi, Mastrolilli, and
  Gambardella]{vercesiGenerationMetricTSP2023}
Vercesi, E., Gualandi, S., Mastrolilli, M., and Gambardella, L.~M.
\newblock On the generation of metric {{TSP}} instances with a large
  integrality gap by branch-and-cut.
\newblock \emph{Mathematical Programming Computation}, 15\penalty0
  (2):\penalty0 389--416, June 2023.
\newblock ISSN 1867-2957.
\newblock \doi{10.1007/s12532-023-00235-7}.

\bibitem[Xu et~al.(2015)Xu, Wang, Chen, and Li]{Xu2015EmpiricalEO}
Xu, B., Wang, N., Chen, T., and Li, M.
\newblock Empirical evaluation of rectified activations in convolutional
  network.
\newblock \emph{ArXiv}, abs/1505.00853, 2015.
\newblock URL \url{https://api.semanticscholar.org/CorpusID:14083350}.

\bibitem[Yilmaz \& {Yorke-Smith}(2021)Yilmaz and
  {Yorke-Smith}]{yilmazStudyLearningSearch2021}
Yilmaz, K. and {Yorke-Smith}, N.
\newblock A {{Study}} of {{Learning Search Approximation}} in {{Mixed Integer
  Branch}} and {{Bound}}: {{Node Selection}} in {{SCIP}}.
\newblock \emph{AI}, 2\penalty0 (2):\penalty0 150--178, April 2021.
\newblock ISSN 2673-2688.
\newblock \doi{10.3390/ai2020010}.

\end{thebibliography}
\bibliographystyle{neurips2024}

\clearpage

\appendix
\section{TSP-as-MILP Formulation}\label{sec:tsp-as-milp}
In general, due to the fact that TSP is amongst the most studied problems in discrete optimization, we can expect existing mixed-integer programming systems to have rich heuristics that provide a strong baseline for our method.
Mathematically, we choose the Miller–Tucker–Zemlin (MTZ) formulation~\citep{MTZ}:
\begin{subequations}
\begin{alignat*}{5}
    &\!\min_x         &\quad&\sum^n_{i=1}\sum^n_{j\neq i,j=1} c_{ij}x_{ij}&&\\
    &\text{subject to}&      &\sum^n_{j=1,i\neq j}x_{ij} = 1&&\forall i=1,\dots,n\\
    &                 &      &\sum^n_{i=1,i\neq j}x_{ij} = 1&&\forall j=1,\dots,n\\
    &                 &      &u_1-u_j +(n-1)x_{ij} \leq n-2 &\quad&2\leq i\neq j\leq n\\
    &                 &      &2 \leq u_i\leq n &&2\leq i\leq n\\
    &                 &      &u_i\in\mathbb{Z},x_{ij}\in\{0,1\}&&
\end{alignat*}
\end{subequations}

Effectively this formulation keeps two buffers: one being the actual $(i,j)$-edges travelled $x_{ij}$, the other being a node-order variable $u_i$ that makes sure that $u_i<u_j$ if $i$ is visited before $j$.
There are alternative formulations, such as the Dantzig–Fulkerson–Johnson (DFJ) formulation, which are used in modern purpose-built TSP solvers, but those are less useful for general problem generation:
The MTZ formulation essentially relaxes the edge-assignments and order constraints, which then are branch-and-bounded into hard assignments during the solving process.
This is different to DFJ, which instead relaxes the ``has to pass through all nodes'' constraint.
DFJ allows for subtours (\eg only contain node $A,B,C$ but not $D,E$) which then get slowly eliminated via the on-the-fly generation of additional constraints.
To generate these constraints one needs specialised row-generators which, while very powerful from an optimization point-of-view, make the algorithm less general as a custom row-generator has to intervene into every single node.
However, in our usecase we also do not really care about the ultimate performance of individual algorithms as the reinforcement learner only looks for improvements to the existing node selections.
This means that as long as the degree of improvement can be adequately judged, we do not need state-of-the-art solver implementations to give the learner a meaningful improvement signal.

\section{Uncapacitated facility location Problem}\label{sec:uncapFL}
Mathmatically, the uncapacitated facility location problem can be seen as sending a product $z_{ij}$ from facility $i$ to consumer $j$ with cost $c_{ij}$ and demand $d_j$. One can only send from $i$ to $j$ if facility $i$ was built in the first place, which incurs cost $f_i$.
The overall problem therefore is
\begin{subequations}
\begin{alignat*}{5}
    &\!\min_x         &\quad&\sum^n_{i=1}\sum^m_{i=1}c_{ij}d_jz_{ij} +\sum^n_{i=0}f_ix_i&&\\
    &\text{subject to}&      &\sum^n_{j=1,i\neq j}z_{ij} = 1&&\forall i=1,\dots,m\\
    &                 &      &\sum^n_{i=1,i\neq j}z_{ij} \leq Mx_i&&\forall j=1,\dots,n\\
    &                 &      &z_{ij} \in \{0,1\}&\quad&\forall i,j=1,\dots,n\\
    &                 &      &x_{i} \in \{0,1\}&\quad&\forall i=1,\dots,n\\
\end{alignat*}
\end{subequations}
where $M$ is a suitably large constant representing the infinite-capacity one has when constructing $x_i=1$.
One can always choose $M\geq m$ since that, for the purposes of the polytop is equivalent to setting $M$ to literal infinity.
This is also sometimes referred to as the ``big $M$'' method.

The instance generator by \cite{Kochetov2005} works by setting $n=m=100$ and setting all opening costs at $3000$.
Every city has 10 ``cheap'' connections sampled from $\{0,1,2,3,4\}$ and the rest have cost $3000$, which represents infinity (\ie also invoking the big $M$ method).


\section{Features}
\label{sec:features}

Table~\ref{tab:features} lists the features used on every individual node. The features split into two different types: One being ``model'' features, the other being ``node'' features.
Model features describe the state of the entire model at the currently explored node, while node features are specific to the yet-to-be-solved added node.
We aim to normalize all features with respect to problem size, as \eg just giving the lower-bound to a problem is prone to numerical domain shifts.
For instance a problem with objective $c^Tx, x\in P$ is inherently the same from a solver point-of-view as a problem $10c^Tx, x\in P$, but would give different lower-bounds.
Since NNs are generally nonlinear estimators, we need to make sure such changes do not induce huge distribution shifts.
We also clamp the feature values between $[-10,10]$ which represent ``infinite'' values, which can occur, for example in the optimality gap.
Last but not least, we standardize features using empirical mean and standard deviation.
\begin{table*}[!htb]
    \fontsize{7}{8}\selectfont
    \centering
    \caption{Features used per individual node.}
    \begin{tabular}{cll}
        \toprule
        \multirow{7}{*}{\rotatebox[origin=c]{90}{model features}}&Number of cuts applied& normalized by total number of constraints \\
        &Number of separation rounds\\
        &optimality gap\\
        &lp iterations\\
        &mean integrality gap\\
        &percentage of variables already integral\\
        &histogram of fractional part of variables & 10 evenly sized buckets\\
        \midrule
        \multirow{2}{*}{\rotatebox[origin=c]{90}{node features}}&\\
        & depth of node& normalized by total number of nodes\\
        & node lowerbound& normalized by min of primal and dual bound\\
        & node estimate& normalized by min of primal and dual bound\\\\
        \bottomrule
    \end{tabular}
    
    \label{tab:features}
\end{table*}
These features are inspired by prior work, such as \cite{labassiLearningCompareNodes,yilmazStudyLearningSearch2021}, but adapted to the fact that we do not need \eg explicit entries for the left or right child's optimality gap, as these (and more general K-step versions of these) can be handled by the GNN.

Further, to make batching tractable, we aim to have constant size features. This is different from \eg \cite{labassiLearningCompareNodes}, who utilize flexibly sized graphs to represent each node.
The upside of this approach is that certain connections between variables and constraints may become more apparent, with the downside being the increased complexity of batching these structures and large amounts of nodes used.
This isn't a problem for them, as they only consider pairwise comparisons between nodes, rather than the entire branch-and-bound graph, but for us would induce a great deal of complexity and computational overhead, especially in the larger instances.
For this reason, we represent flexibly sized inputs, such as the values of variables, as histograms: i.e., instead of having $k$ nodes for $k$ variables and wiring them together, we produce once distribution of variable values with 10-buckets, and feed this into the network.
This looses a bit of detail in the representation, but allows us to scale to much larger instances than ordinarily possible.

In general, these features are not optimized, and we would expect significant improvements from more well-tuned features. 
Extracting generally meaningful features from branch-and-bound is a nontrivial task and is left as a task for future work.

\section{Architecture}\label{sec:architecture}
Our network consists of two subsystems:
First, we have the feature embedder that transforms the raw features into embeddings, without considering other nodes this network consists of one linear layer $|d_{features}|\to |d_{model}|$ with LeakyReLU~\citep{Xu2015EmpiricalEO} activation followed by two $|d_{model}|\to |d_{model}|$ linear layers (activated by LeakyReLU) with skip connections. We finally normalize the outputs using a Layernorm~\citep{ba2016layer} \emph{without} trainable parameters (\ie just shifting and scaling the feature dimension to a normal distribution).

Second, we consider the GNN model, whose objective is the aggregation across nodes according to the tree topology.
This consists of a single LeakyReLU activated layer with skip-connections.
We use ReZero~\citep{Bachlechner2020ReZeroIA} initialization to improve the convergence properties of the network.
Both the weight and value heads are simple linear projections from the embedding space.
Following the guidance in \citep{andrychowiczWhatMattersOnPolicy2020}, we make sure the value and weight networks are independent by detaching the value head's gradient from the embedding network.

For this work we choose $|d_{model}|=512$, but we did not find significant differences between different model sizes past a width of 256.
For training we use AdamW~\cite{Loshchilov2017FixingWD} with a standard learning rate of $3\cdot 10^{-4}$ and default PPO parameters.

\section{Necessity of GNN}\label{sec:ablateGNN}

In addition to the tests done above, we also investigated running the model without a GNN: 
We found that when removing the GNN, the model tended to become very noisy and produce unreproducible experiments.
Considering only the cases where the GNN-free model did well, we still found the model needed roughly 30\% more nodes than the SCIP or our model with a GNN.
More importantly, we notice the GNN-free model diverges during training: starting with a reward of roughly zero, the model diverges down to a reward of $\approx-0.2$, which amounts to a score roughly 20\% worse than SCIP.
We therefore conclude that, at least for our architecture, the GNN is necessary for both robustness and performance.

\section{Theoretical Derivation}\label{sec:theoryNodeSelect}

\begin{figure}[htb]
    \centering
    \includegraphics[width=0.75\linewidth]{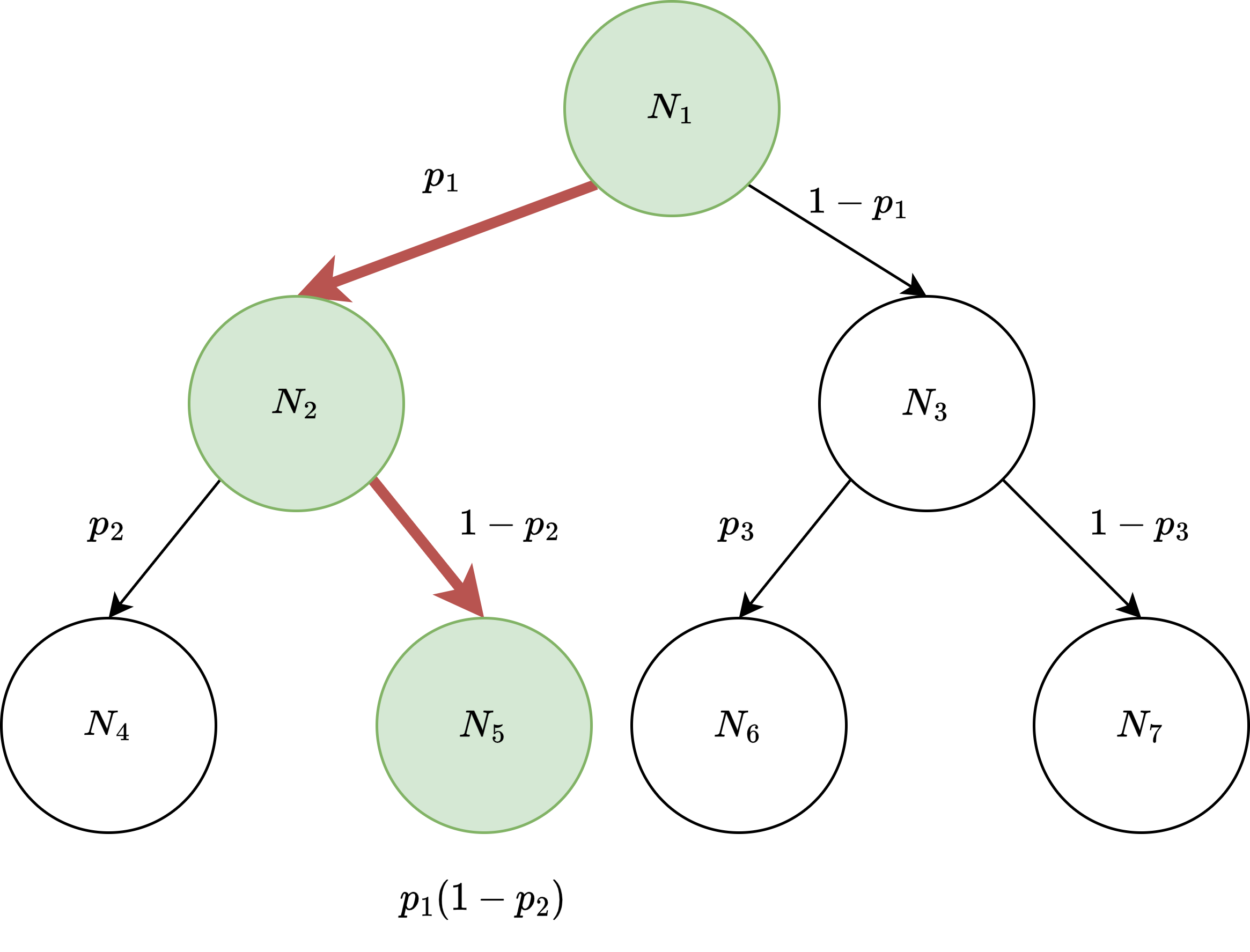}
    \caption{Naive approach using recursive selection. The probabilities are computed based on which ``fork'' of the tree is traveled. Sampling this can be done by sampling left or right based on $p_i$}
    \label{fig:recursiveSelection}
\end{figure}

One naive way of parameterizing actions is selecting probabilistically by training the probability of going to the left or right child at any node.
This effectively gives a hierarchical Bernoulli description of finding a path in the tree.
A path is simply a sequence of left (``zero direction'') and right (``one direction'') choices, with the total probability of observing a path being the likelihood of observing a specific chain
\begin{equation*}
    p(\text{leaf}) = \prod_{i\in \text{P(root,leaf)}}p_i,
\end{equation*}
with $p_i$ being the i'th decision in the graph (see Fig~\ref{fig:recursiveSelection}).
Node selection can now be phrased as the likelihood of a random (weighted) walk landing at the specified node.
However, using this parametrization directly would yield a significant number of problems:

Removing illegal paths from the model is quite challenging to do with sufficiently high performance. If a node gets pruned or fully solved, the probability to select this node has to be set to zero.
This information has to be propagated upward the tree to make sure that the selection never reaches a ``dead-end''.
This is possible, but turns out to be rather expensive to compute since it needs to evaluate the entire tree from leaves to root.

From a theoretical point of view, the hierarchical Bernoulli representation also has a strong prior towards making selections close to the root, as the probability of reaching a node a depth $K$ consists of $p_1\cdot p_2\cdot \dots \cdot p_K$ selections. Since all $p_i$ are set to an uninformative prior $0<p_i< 1$, the model at initialization has an increased likelihood to select higher nodes in the tree. 
Considering that classical methods specifically mix plunging heuristics into their selection to get increased depth exploration (see Sec.~\ref{sec:RelatedWork}), this ``breadth over depth'' initialization is expected to give low performance.
Therefore, one would need to find a proper initialization that allows the model to \eg uniformly select nodes independent of depth to get good initialization.

The fact that multiple sampling sites exist in this method also poses issues with RL, as small errors early in the tree can have catastrophic effects later.
This means the tree based model is a significantly higher variance estimator than the estimator we propose in Sec~\ref{subsec:MethodRL}, which yields much worse optimization characteristics.
It is well established in existing deep learning (\eg \citep{Kingma2013AutoEncodingVB}) and general Bayesian Inference (\eg \citep{Robert2021RaoBlackwellizationIT,f47e9344-75dc-3782-a811-9e51a952e8bf}) literature, that isolating or removing sampling sites can lead to lower variance, and therefore much more performant, statistical estimators.

Using those insights, we can now rewrite this naive approach into the one we propose in Sec.~\ref{subsec:MethodRL}:
First, instead of sampling just one path, we can sample \emph{all} possible paths and compute the likelihood for each.
The result will be a probability $p$ at every possible leaf.
If one parameterizes this as a logarithmic probability, we can write 
\begin{equation}
    \log p_{\text{leaf}} = \sum_{i\in P(root,\text{leaf})} \log p_i.
\end{equation}
We can further assume that the $\log p$'s are given as \emph{unnormalized logits} $W'$ and only normalize them at the end such that to likelihood of all leaves together is $1$.

This already gives an increase in performance, as the paths can be computed in parallel \emph{without} needing iterated random sampling, which tends to be a very expensive operation.
One also has fewer discrete choices, meaning fewer backpropagation graph-breaks, which improves overall stability.
The change to unnormalized logits acts essentially as a reparametrization trick (for other examples of reparametrization, see \citep{Kingma2013AutoEncodingVB}).

If one considers unnormalized log-probabilities, which have a range $(-\infty,\infty)$, this scheme becomes very similar to the one we ended up using in Sec.~\ref{subsec:MethodRL}.
Let $f(\text{node},\text{tree})$ denote the unnormalized probabilities such that $p(\text{node},\text{tree}) = \frac{f(\text{node},\text{tree})}{Z}$.
One can compute the likelihood of selecting a specific node as:
\begin{align*}
    p(\text{leaf}|\text{tree}) &= \frac{f(\text{leaf},\text{tree})}{\sum_{c\in\mathscr{C}}f(c,\text{tree})}&&\text{normalization}\\
    &=\frac{\exp(\log (f(\text{leaf},\text{tree}))}{\sum_{c\in\mathscr{C}}\exp(\log(f(c,\text{tree})}&&\text{$\log f$ are unnormalized logits}\\
    &=\frac{\exp(W'(\text{leaf}))}{\sum_{c\in\mathscr{C}}\exp(W'(c))}&&\text{definition W' as unnormalized logit}\\
    &=\operatorname{softmax}_{\text{leaf}}\{W'(c) | \forall c\in\mathscr{C}\} &&\text{definition softmax},
\end{align*}
where $\operatorname{softmax}_{\text{leaf}}$ refers to the softmax-entry associated with $\text{leaf}$.
The resulting decomposition is, while not the same, equivalent to the original Bernoulli decomposition (in the sense that for any Bernoulli decomposition there exists a softmax-based decomposition). 
Beyond the massive reduction in sampling sites from $O(depth)$ to a single sampling site, phrasing the problem using unnormlized-$\log p$ representations gives rise to additional optimizations:

Firstly, to achieve uniform likelihood for all nodes at the beginning of training, we simply have to initialize $W(n)=0$ for all nodes, which can be done by setting the output weights and biases to zero (in practice we set bias to zero and weights to a small random value).

Secondly to account for pruned or solved nodes, we do not need to change the unnormalized probabilities.
Instead, we simply only sample the paths associated with selectable nodes, as we know the likelihood of selecting a pruned or solved node is zero.
This means we can restrict ourselves to only evaluating the candidate paths $\mathscr{C}$, rather than all paths (see Eq.~\ref{eq:policy}).

The last modification we made in our likelihood computation (see Eq.~\ref{eq:Weight}), is to normalize the weights based on depth.
This is done mainly for numerical stability, but in our Bayesian framework is equivalent to computing the softmax-normalization Eq.~\ref{eq:policy} using weights inversely proportional to depth.
The resulting algorithm is much more stable and performant than the naive parametrization.
In early implementations we used the recursive-sampling approach to compute likelihoods, but the resulting scheme did not manage to reach above random results on the training set, presumably due to the higher computational burden and worse initialization.
There are also other advantages to our parametrization, such as the fact that one can easily include a sampling temperatur $\tau$ into the process
$$\text{node}\sim\operatorname{softmax}\{W'(c) / \tau | \forall c\in\mathscr{C}\}$$
which can be nice to tune the model more into the direction ``exploitation'' during testing.
For our later benchmarking (Sec~\ref{sec:Experiments}) we keep $\tau=1$ (\ie no temperature), as it would introduce a further tunable hyperparameter, but this could be interesting for future work.

This construction has interesting parallels to Monte-Carlo Tree Search (MCTS)~\cite{Silver2017MasteringCA} based reinforcement learning methods.
The difference in our method is that we consider iterative updates to a policy distribution $\pi$ rather than a Q-function, and therefore can be seen as a policy-based tree search scheme.
We leave a more thorough investigation of this to future work.

{
\section{Additional metrics}
\begin{table*}[t]
    \small
    \centering
    \caption{Performance across benchmarks (the policy only saw TSP instances during training). The 5\si{min} runs use the same model, evaluated for the first 650 nodes, and processed according to Sec.~\ref{subsec:Experiments:Baselines}.
    }
    \vspace{-2mm}
    \resizebox{\textwidth}{!}{
    \begin{tabular}{lcccccc}\toprule
         Benchmark                &Reward& Utility & Utility/Node & Win-rate & geo-mean Ours & geo-mean SCIP \\\midrule
         TSPLIB~\citep{TSPLib}     &0.184 & 0.030   &0.193         & 0.50                     & 0.931                          & 0.957 \\
         UFLP~\citep{Kochetov2005} &0.078 & 0.093   & -0.064       & 0.636                    & 0.491                          & 0.520 \\
         MINLPLib~\citep{MINLPLIB} &0.487 & 0.000   &0.114         & 0.852                    & 28.783                         & 31.185\\
         MIPLIB~\citep{MIPLIB}     &0.140 & -0.013  &0.208         & 0.769                    & 545.879                        & 848.628\\
      TSPLIB@5min &0.192 & 0.056   &0.336         & 0.600                    & 1.615                          & 2.000 \\
     MINLPlib@5min&0.486 & -0.012  &0.078         & 0.840                    & 17.409                         & 20.460\\
     MINLPlib@5min&0.150 &-0.075   &0.113         & 0.671                    & 66.861                         & 106.400\\
         \bottomrule
    \end{tabular}
    }
    \label{tab:otherTests}
\end{table*}
Aside from the previously discussed Reward in section~\ref{sec:RewardDefinition}, we also considered two different ways of measuring the quality of our method compared to SCIP. While both of these have theoretical problems, we still find them interesting to consider as they act as additional reference points for the behavior of the policies.

\textbf{Utility} defines the \emph{difference} between both methods normalized using the maximum of both gaps:
\begin{equation}
    \operatorname{Utility} = \left(\frac{\text{\textit{gap(scip)}}-\text{\textit{gap(node selector)}}}{\max\left(\text{\textit{gap(node selector)}},\text{\textit{gap(scip)}}\right)}\right).
\end{equation}
The reason we do not use this as a reward measure is because we empirically found it to produce worse models. 
This is presumably because some of the negative attributes of our reward, e.g., the asymmetry of the reward signal, lead to more robust policies.
In addition, the utility metric gives erroneous measurements when both models hit zero optimality gap. 
This is because utility implicitly defines $\frac{0}{0}=0$, rather than reward, which defines it as $\frac{0}{0}=1$.
In some sense the utility measurement is accurate, in that our method does not improve upon the baseline.
On the other hand, our method is already provably optimal as soon as it reaches a gap of $0\%$.
In general, utility compresses the differences more than reward which may or may not be beneficial in practice.

\textbf{Utility per Node} normalizes Utility by the number of nodes used during exploration:
\begin{subequations}
\begin{align}
    &\operatorname{Utility/Node} = \left(\frac{\operatorname{scip}-\operatorname{selector}}{\max\left(\operatorname{selector},\operatorname{scip}\right)}\right),
\end{align}
\end{subequations}
where $\operatorname{selector} = \frac{\text{\textit{gap(node selector)}}}{\text{\textit{nodes(node selector)}}}$ and $\operatorname{scip} = \frac{\text{\textit{gap(scip)}}}{\text{\textit{nodes(scip)}}}$.
The per-node utility gives a proxy for the total amount of ``work'' done by each method.
However, it ignores the individual node costs, as solving the different LPs may take different amounts of resources (a model with higher ``utility/node'' is not necessarily more efficient as our learner might pick cheap but lower expected utility nodes on purpose). 
Further, the metric is underdefined: comparing two ratios, a method may become better by increasing the number of nodes processed, but keeping the achieved gap constant.
In practice the number of nodes processed by our node selector is dominated by the implementation rather than the node choices, meaning we can assume it is invariant to changes in policy.
Another downside arises if both methods reach zero optimality gap, the resulting efficiency will also be zero regardless of how many nodes we processed.
As our method tend to reach optimality much faster (see Sec.~\ref{sec:Experiments} and Appendix~\ref{sec:fullResults}), all utility/node results can be seen as a lower-bound for the actual efficiency.

}

{
\section{Comparisons of Runtime to completion}
\begin{table}[]
    \centering
    \begin{tabular}{lccc}\toprule
        Method &Runtime shifted geometric mean & Nodes Processed geometric mean\\ \midrule
        SCIP & 1016.61\si{s} & 4122.67\\
        Ours & 1003.03\si{s} & 2735.33\\\bottomrule
    \end{tabular}
    \caption{The shifted geometric mean (shifted by +10) of runtime and number of nodes processed.}
    \label{tab:runtime_30}
\end{table}

We also compare the runtime of our selector against the SCIP baseline on TSPLIB. For this we set a maximum time limit of 30\si{min} and run both methods. It is worth noting that due to the inherent difficulty of TSPLIB, a significant number of problems are still unsolved after 30\si{min}, in which case these models are simply assigned the maximum time as their overall completion time.
Our method outperforms the baseline, however due to the fact that our method aims to improve the quality of early selections in particular, the amount of improvement is rather small.
Further, our method is fundamentally undertrained to adequately deal with longer time horizons and we would assume the effect of our method is larger the longer the model is trained.
Despite this being a worst-case evaluation scenario, our method manages to outperform the baseline in both runtime and number of nodes needed.

Interestingly, we also manage to reach this superior performance while utilizing only about half the number of nodes. This discrepancy cannot be explained by just considering the slowdown due to our policy needing to be evaluated, as, while it is a significant overhead due to the python implementation, it is very far from being large enough to justify a 50\% slowdown over the course of 30\si{min}. 
This also shows more generally that even if one restricts themselves to only processing the beginning of the branch-and-bound tree, one can reach superior performance to the standard node-selection schemes.
}
\section{Full Results}\label{sec:fullResults}
The following two sections contain the per-instance results on the two ``named'' benchmarks TSPLIB~\citep{TSPLib} and MINLPLIB~\citep{MINLPLIB}.
We test against the strong SCIP 8.0.4 baseline. 
Due to compatibility issues, we decided not to test against \citep{labassiLearningCompareNodes} or \citep{heLearningSearchBranch}: These methods were trained against older versions of SCIP, which not only made running them challenging, but also would not give valid comparisons as we cannot properly account for changes between SCIP versions.
\cite{labassiLearningCompareNodes} specifically relies on changes to the SCIP interface, which makes porting to SCIP 8.0.4 intractable.
In general, this shouldn't matter too much, as SCIP is still demonstrably the state-of-the-art non-commercial mixed-integer solver, which frequently outperforms even closed-source commercial solvers (see \cite{DecisonTreeOptimization} for thorough benchmarks against other solvers), meaning outperforming SCIP can be seen as outperforming the state-of-the-art.
\subsubsection{Kochetov-UFLP}
\label{sec:exp:UFLP}

To demonstrate the generalizability of the learned heuristics, we test our method on the Uncapacitated Facility Location Problem (see Appendix~\ref{sec:uncapFL}) \emph{without further finetuning}, i.e., we only train on TSP instances and never show the algorithm any other linear or nonlinear problem.
For testing, we generate 1000 instances using the well-known problem generator by \cite{Kochetov2005}, which was designed to have large optimality gaps, making these problems particularly challenging.

Our method performs very similar to the highly optimized baseline, despite never having seen the UFL problem, see Table~\ref{tab:otherTests}.
We argue that this is specifically because our method relies on tree-wide behaviour, rather than individual features to make decisions.
We further hypothesize that the reason for the advantage over the baseline being so small is due to the fact that UFLP consists of ``adversarial examples'' to the branch-and-bound method where cuts have reduced effectiveness.
This means clever node-selection strategies have limited impact on overall performance.

An interesting aspect is that our method processes more nodes than the baseline, which also leads to the loss in node-efficiency.
This implies that our method selects significantly easier nodes, as ordinarily our solver is slower just due to the additional overhead.
Considering that this benchmark was specifically designed to produce high optimality gaps, it makes sense that our solver favours node quantity over quality, which is an interesting emergent behaviour of our solver.

\subsection{TSPLIB results}\label{sec:Tsplib-res}
\begin{table*}[!h]
    \centering
    \caption{Results on TSPLIB~\citep{TSPLib} after 45s runtime. Note that we filter out problems in which less than 5 nodes were explored as those problems cannot gain meaningful advantages even with perfect node selection. ``Name'' refers to the instances name, ``Gap Base/Ours'' corresponds to the optimization gap achieved by the baseline and our method respectively (lower is better), ``Nodes Base/Ours'' to the number of explored Nodes by each method, and ``Reward'', ``Utility'' and ``Utility Node'' to the different performance measures as described in Section~\ref{sec:Experiments}.}
    \resizebox{0.75\textwidth}{!}{
\begin{tabular}{lrrrrrrr}
\toprule
     Name &  Gap Ours &  Gap Base &  Nodes Ours &  Nodes Base &  Reward &  Utility &  Utility/Node \\
\midrule
    att48 &     0.287 &     0.286 &        1086 &        2670 &  -0.002 &   -0.002 &         0.593 \\
   bayg29 &     0.000 &     0.000 &        2317 &        7201 &   1.000 &    0.000 &         0.000 \\
   bays29 &     0.000 &     0.036 &       11351 &       10150 &   1.000 &    1.000 &         0.997 \\
 berlin52 &     0.000 &     0.000 &         777 &        1634 &   1.000 &    0.000 &         0.000 \\
  bier127 &     2.795 &     2.777 &          23 &          25 &  -0.007 &   -0.007 &         0.074 \\
 brazil58 &     0.328 &     0.644 &        1432 &        2182 &   0.491 &    0.491 &         0.666 \\
  burma14 &     0.000 &     0.000 &          96 &          65 &   1.000 &    0.000 &         0.000 \\
\bottomrule
\end{tabular}

}
    \label{tab:ExplorationData}
\end{table*}

\clearpage
\onecolumn
\subsection{MIPLIB results}\label{sec:MIPLIB}
{

\fontsize{7}{8}\selectfont
\begin{longtable}{lrrrrrrr}
\caption{Results on MIPLIB~\citep{MIPLIB} after 45s runtime. Note that we filter out problems in which less than 5 nodes were explored as those problems cannot gain meaningful advantages even with perfect node selection. ``Name'' refers to the instances name, ``Gap Base/Ours'' corresponds to the optimization gap achieved by the baseline and our method respectively (lower is better), ``Nodes Base/Ours'' to the number of explored Nodes by each method, and ``Reward'', ``Utility'' and ``Utility Node'' to the different performance measures as described in Section~\ref{sec:Experiments}. Note that all results where achieved with a policy only trained on TSP instances}\\ 
\toprule
               Name &  Gap Ours &  Gap Base &  Nodes Ours &  Nodes Base &  Reward &  Utility &  Utility/Node \\
\midrule
\endfirsthead

\toprule
               Name &  Gap Ours &  Gap Base &  Nodes Ours &  Nodes Base &  Reward &  Utility &  Utility/Node \\
\midrule
\endhead
\midrule
\multicolumn{8}{r}{{Continued on next page}} \\
\midrule
\endfoot

\bottomrule
\endlastfoot
            30n20b8 &     2.662 &  $\infty$ &         147 &         301 &   1.000 &    1.000 &         1.000 \\
             50v-10 &     0.101 &     0.113 &         303 &        1094 &   0.103 &    0.103 &         0.752 \\
          CMS750\_4 &     0.100 &     0.072 &          68 &         281 &  -0.389 &   -0.280 &         0.664 \\
              air05 &     0.000 &     0.000 &         248 &         523 &   1.000 &    0.000 &         0.000 \\
        assign1-5-8 &     0.085 &     0.087 &       17466 &       23589 &   0.030 &    0.030 &         0.282 \\
        binkar10\_1 &     0.000 &     0.000 &        2843 &        2270 &   1.000 &    0.000 &         0.000 \\
           blp-ic98 &     0.127 &     0.127 &          26 &          43 &   0.001 &    0.001 &         0.396 \\
           bnatt400 &  $\infty$ &  $\infty$ &         547 &        1568 &   0.000 &    0.000 &         0.651 \\
\end{longtable}
}

\subsection{MINLPLIB results}\label{sec:MINLP}
{

\fontsize{7}{8}\selectfont
\begin{longtable}{lrrrrrrr}
\caption{Results on MINLPLIB~\citep{MINLPLIB} after 45s runtime. Note that we filter out problems in which less than 5 nodes were explored as those problems cannot gain meaningful advantages even with perfect node selection. ``Name'' refers to the instances name, ``Gap Base/Ours'' corresponds to the optimization gap achieved by the baseline and our method respectively (lower is better), ``Nodes Base/Ours'' to the number of explored Nodes by each method, and ``Reward'', ``Utility'' and ``Utility Node'' to the different performance measures as described in Section~\ref{sec:Experiments}. For all three measures, higher is better.}\\ 
\toprule
                          Name &  Gap Ours &  Gap Base &  Nodes Ours &  Nodes Base &  Reward &  Utility &  Utility/Node \\
\midrule
\endfirsthead

\toprule
                          Name &  Gap Ours &  Gap Base &  Nodes Ours &  Nodes Base &  Reward &  Utility &  Utility/Node \\
\midrule
\endhead
\midrule
\multicolumn{8}{r}{{Continued on next page}} \\
\midrule
\endfoot

\bottomrule
\endlastfoot
                 ball\_mk4\_05 &     0.000 &     0.000 &        1819 &        1869 &   1.000 &    0.000 &         0.000 \\
                 ball\_mk4\_10 &  $\infty$ &  $\infty$ &       31684 &       37656 &   0.000 &    0.000 &         0.159 \\
                 ball\_mk4\_15 &  $\infty$ &  $\infty$ &        1773 &        2415 &   0.000 &    0.000 &         0.266 \\
                    bayes2\_20 &  $\infty$ &  $\infty$ &        3171 &        2719 &   0.000 &    0.000 &        -0.143 \\
                    bayes2\_30 &  $\infty$ &  $\infty$ &        4462 &        4992 &   0.000 &    0.000 &         0.106 \\
                    bayes2\_50 &  $\infty$ &  $\infty$ &        2934 &        2530 &   0.000 &    0.000 &        -0.138 \\
                      blend029 &     0.000 &     0.000 &         812 &         804 &   1.000 &    0.000 &         0.000 \\
                      blend146 &     0.097 &     0.105 &       12390 &       18066 &   0.075 &    0.075 &         0.365 \\
\end{longtable}
}


\end{document}